\documentclass{IJCAS}

\usepackage{url}
\usepackage{array,tabularx}
\usepackage{multicol,multirow} 
\usepackage{amsmath,amssymb,amsfonts}
\usepackage{algorithmic}
\usepackage{graphicx}
\usepackage{diagbox}
\usepackage[dvipsnames]{xcolor}

\setarticlestartpagenumber{1}

\allowdisplaybreaks

\begin{document}
\title{Leveraging Spatial Attention and Edge Context for Optimized Feature Selection in Visual Localization}

\author{Nanda Febri Istighfarin* and HyungGi Jo*}

\begin{abstract}
Visual localization determines an agent's precise position and orientation within an environment using visual data. It has become a critical task in the field of robotics, particularly in applications such as autonomous navigation. This is due to the ability to determine an agent’s pose using cost-effective sensors such as RGB cameras. Recent methods in visual localization employ scene coordinate regression to determine the agent’s pose. However, these methods face challenges as they attempt to regress 2D-3D correspondences across the entire image region, despite not all regions providing useful information. To address this issue, we introduce an attention network that selectively targets informative regions of the image. Using this network, we identify the highest-scoring features to improve the feature selection process and combine the result with edge detection. This integration ensures that the features chosen for the training buffer are located within robust regions, thereby improving 2D-3D correspondence and overall localization performance. Our approach was tested on the outdoor benchmark dataset, demonstrating superior results compared to previous methods.
\end{abstract}


\maketitle

\makeAuthorInformation{
*Jeonbuk National University
}

\runningtitle{2024}{Nanda Febri Istighfarin, HyungGi Jo}{Leveraging Spatial Attention and Edge Context for Optimized Feature Selection in Visual Localization}{xxx}{xxxx}{x}

\section{INTRODUCTION}
Visual localization is gaining popularity in research fields such as virtual reality and autonomous vehicles. The initial method used in visual localization is the feature matching-based approach \cite{sattler2016efficient,sarlin2019coarse}. The feature matching-based method is known for its high localization accuracy. This technique depends on matching the 2D points in a query image to 3D points in scene coordinates. The 3D map itself can be retrieved using COLMAP \cite{schonberger2016structure} or Visual SfM \cite{wu2011visualsfm}. However, the use of a pre-existing 3D map in the pipeline can be costly and inefficient due to the significant memory and time required to create these maps.

Another method for visual localization involves deep learning, where neural networks are used either to directly regress the pose or to estimate it through an additional step. Absolute Pose Regression (APR) such as \cite{kendall2015posenet,kendall2017geometric} employs neural networks to directly infer the camera pose from a query image. Despite its direct approach, APR often fails to produce accurate localization results.

Alternatively, Scene Coordinate Regression (SCR) \cite{brachmann2018learning} uses neural networks to establish 2D-3D correspondences from a query image to an implicit map representation. These correspondences are then input into a pose solver like the RANSAC solver to determine the camera pose. SCR typically achieves better localization results than APR, but it suffers from long training times. To address this issue approaches like \cite{dong2022visual} and \cite{brachmann2023accelerated} involve dividing the network into a scene-agnostic backbone and a scene-specific head regressor, which speeds up the mapping process. Furthermore, in ACE \cite{brachmann2023accelerated}, they enhance the training efficiency through gradient decorrelation by utilizing a training buffer that includes diverse mappings of views of images. This approach achieves faster convergence and a robust result. However, the selection of features for the training buffer is random, sometimes including less informative parts of the image.

In this paper, we improve feature selection by using the spatial information of image patches, choosing only the most relevant and highest-scoring features for the training buffer. We also include the edge detector along our pipeline to ensure the relevant features are placed in the robust region. This work extends the foundational research in a Master's thesis \cite{istighfarin2024}. Our approach also surpasses the localization result of its baseline, ACE. Our contributions are summarized as follows:

\begin{enumerate}
    \item We employ a simple attention network to leverage spatial information between image patches upon the ACE \cite{brachmann2023accelerated} network, ensuring that only informative features are selected.
    \item We integrated the edge detector with the simple attention network in our sampling module to ensure that the sampled features are located in robust regions.
    \item We show that our sampling method maintains quick mapping times and efficient mapping sizes, despite using an attention network and edge detector in the pipeline.
    \item We evaluate our proposed method using benchmark datasets, confirming its superiority over previous approaches.  
\end{enumerate}

\section{RELATED WORK}
\subsection{Feature Matching-based Method}
The feature matching-based method utilizes the 2D-3D correspondences of a given query image to a 3D points map for calculating the camera poses. Tools like COLMAP \cite{schonberger2016structure,schonberger2016pixelwise} and Visual SfM \cite{wu2011visualsfm} are frequently utilized to generate the 3D map. To establish these correspondences, the method involves searching for local features in the images using either a handcrafted feature extractor like SIFT \cite{lowe2004distinctive} or a CNN-based extractor such as R2D2 \cite{revaud2019r2d2}, SuperPoint \cite{detone2018superpoint}, or DISK \cite{tyszkiewicz2020disk}. Consequently, we match these extracted features to the 3D map using either discrete matching or CNN-based techniques like SuperGlue \cite{sarlin2020superglue}. 

Given that matching can be resource-intensive, especially with large datasets, image retrieval techniques \cite{arandjelovic2016netvlad,ge2020self,Berton_CVPR_2022_CosPlace,arshad2023robust,lee2019bag,park2020robust} are used to filter and match only those query images that visually resemble the map images. Once we retrieved the 2D-3D correspondences, we passed these correspondences into the pose solver like the Kabsch\cite{kabsch1976solution}, PnP\cite{lepetit2009ep} algorithm in RANSAC or its differentiable variance\cite{brachmann2017dsac}. While the feature matching-based method yields accurate pose estimations, the requirement for a 3D map makes it a costly approach.

\subsection{Deep learning-based Method}
As its name, the deep learning-based method utilizes a neural network in the localization pipeline. Absolute pose regression (APR) is one of the visual localization methods that directly regresses the camera pose from a given query image through a neural network. Various methods have already been developed in APR including \cite{kendall2015posenet, kendall2016modelling, shavit2021learning}. The camera pose estimated by the APR method is relative to a pre-defined world coordinate system, which differentiates it from the Relative Pose Regression (RPR) method. RPR \cite{melekhov2017relative,balntas2018relocnet,en2018rpnet} calculates the pose between two images, typically comparing a current view with a reference image or between consecutive images. Nonetheless, both APR and RPR tend to produce less accurate results.

There has been further development in deep learning-based methods for visual localization, such as the Scene Coordinate Regression (SCR) method. SCR maps a 2D image position to a 3D position on an implicit map directly through a neural network. Once the 2D-3D correspondences are established, they are inputted into a pose solver to estimate the camera pose. The initial proposed SCR \cite{shotton2013scene,valentin2015exploiting} required RGB-D input, making it a drawback when we use inexpensive sensors only, i.e. RGB camera. However, the newer version, DSAC* \cite{brachmann2021visual}, is able to only use RGB images as the network input. It also achieves impressive results, though it requires about 15 hours of training time on a powerful GPU, which is a significant limitation.

A novel advancement in SCR is the development of an accelerated network, ACE \cite{brachmann2023accelerated}, which significantly reduces training time. Inspired by previous works \cite{dong2022visual}, the SCR network is divided into two parts: a scene-agnostic backbone and a scene-specific head regressor. The scene agnostic backbone is a pre-trained network that is trained on 100 scenes of ScanNet \cite{dai2017scannet} dataset. The backbone later extracts features that fill the training buffer, which is then used to train the scene-specific head network. However, ACE employs random sampling when populating the training buffer with extracted features. We argue that the training buffer should be filled only with the most informative and relevant features to ensure the network effectively learns the 2D-3D correspondences.

In this paper, we enhance feature selection by using spatial information and edge detection to select only the most relevant, highest-scoring features located in robust regions for our training buffer. We also explore whether this approach can improve localization accuracy.

\section{METHODOLOGY}
\subsection{Preliminaries}
Consider a dataset of grayscale images denoted by \( I_G \), which can be partitioned into two distinct subsets: mapping images \( I_M = \{M_1, M_2, \ldots, M_m\} \) and query images \( I_Q = \{Q_1, Q_2, \ldots, Q_n\} \). 

Our primary objective is to predict the 6-DoF camera pose \( h_i \) for each query image \( Q_i \in I_Q \). \( h_i \) is a transformation in the special Euclidean group \( SE(3) \) that maps coordinates from the camera coordinate space \(\mathbf{e}\) to the scene coordinate space \(\mathbf{y}\):

\vspace{-1.5em}
\begin{align}
    \label{eq:objective_h}
    \mathbf{y}_k = h_i\mathbf{e}_k,
\end{align}

where \( k \) denotes the pixel index in the query image \( Q_i \). To estimate \(\mathbf{y}_k\) for the given query image \( Q_i \), we utilize a neural network \(f\) parameterized by weight \(\mathbf{w}\):

\vspace{-1.5em}
\begin{align}
    \label{eq:neural_network}
    \mathbf{y}_k = f(\mathbf{p}_k;\mathbf{w}), \text{where } \mathbf{p}_k = \mathcal{P}(\mathbf{x}_k, Q_i).
\end{align}

\(\mathbf{p}_k\) denotes the pixel patch extracted from position \(\mathbf{x}_k\) in the query image \( Q_i \). Therefore the neural network \(f\) will map a grayscale image into the scene coordinate position, \(\mathbb{R}^{{\mathrm{1}} \times \mathrm{H_p} \times \mathrm{W_p}} \to \mathbb{R}^{\mathrm{3} \times \mathrm{H_p} \times \mathrm{W_p}}\), where \(\mathrm{H_p}\) and \(\mathrm{W_p}\) denotes the patch dimension that set to 81px \cite{brachmann2018learning,brachmann2021visual,brachmann2023accelerated}. Following \cite{brachmann2023accelerated}, there will be no explicit patch extraction in our network. Nevertheless, explicit patch notation is used to enhance a deeper comprehension of the network's operations.

In \cite{dong2022visual, brachmann2023accelerated}, dividing the regression network into a convolutional backbone and a multi-layer perceptron (MLP) scene-specific head regressor has been shown to reduce training duration. Consequently, Eq. \ref{eq:neural_network} is revised as:

\vspace{-1.5em}
\begin{align}
    \label{eq:split_network}
    f(\mathbf{p}_k;\mathbf{w}) = f_\mathrm{H}(\mathfrak{f}_k; \mathbf{w}_\mathrm{H}), \text{with } \mathfrak{f}_k = f_\mathrm{B}(\mathbf{p}_k;\mathbf{w}_\mathrm{B}),
\end{align}

where \(\mathfrak{f}_k\) is a high-dimensional feature with hard-coded dimensionality 512 that is generated by the backbone \(f_\mathrm{B}\) and the scene-specific head regressor \(f_\mathrm{H}\) is the network to predict the scene coordinates. Therefore, the backbone network maps \(f_\mathrm{B}:\mathbb{R}^{{\mathrm{1}} \times \mathrm{H_p} \times \mathrm{W_p}} \to \mathbb{R}^{\mathrm{512} \times \mathrm{H_p} \times \mathrm{W_p}}\) and the scene-specific head regressor network performs \(f_\mathrm{H}:\mathbb{R}^{\mathrm{512}\times \mathrm{H_p} \times \mathrm{W_p}} \to \mathbb{R}^{\mathrm{3}\times \mathrm{H_p} \times \mathrm{W_p}}\).

\begin{figure*}[t!]
\begin{center}
\includegraphics[width=\textwidth]{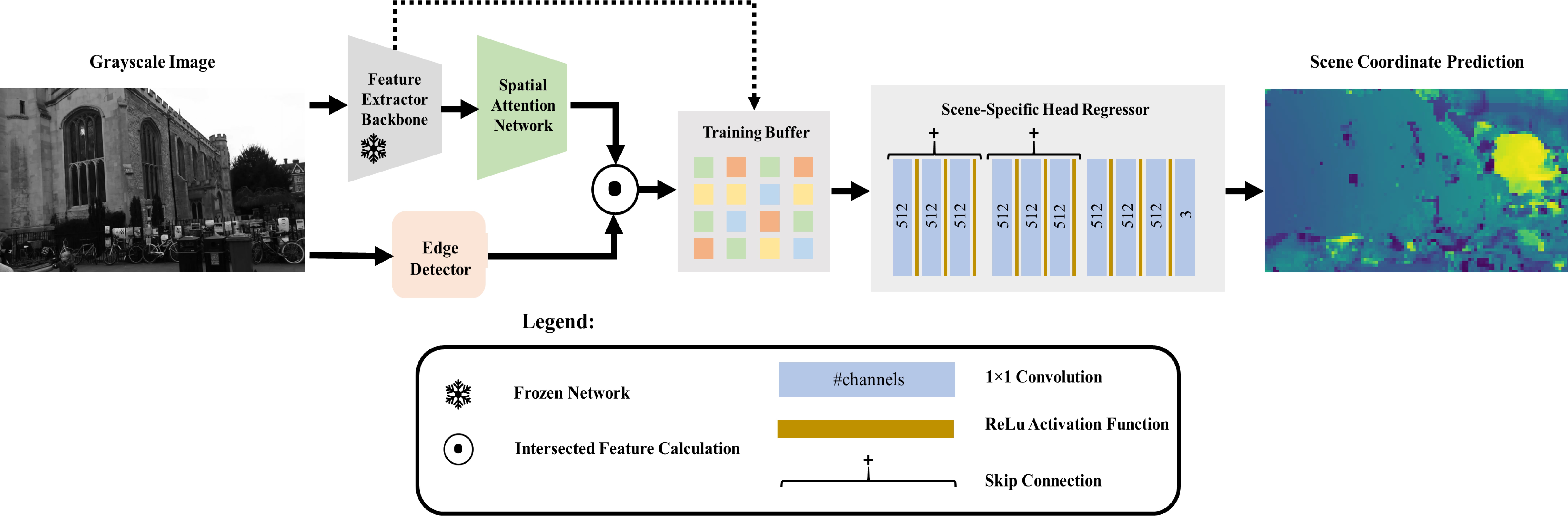}
\caption{\label{fig:framework}Proposed Network Architecture. We integrate the spatial attention network and edge detector results to ensure that only informative features are included in the training buffer.}

\end{center}
\vskip -1.5pc
\end{figure*}

The network \(f\) in Eq. \ref{eq:split_network} then trained on our mapping images \(I_M\) and their ground truth pose \(h_i^*\) by minimizing the re-projection objective function \cite{brachmann2021visual}:

\vspace{-1.5em}
\begin{align}
    \label{eq:reprojection}
    r(\mathbf{x}_k,\mathbf{y}_k,h_i^*) = \|\pi(\mathbf{K}_k{h_i^*}^{-1}\mathbf{y}_k)-\mathbf{x}_k\|,
\end{align}

where \(h_i^*\) is the ground truth camera pose, \(\pi(.)\) represents the process of mapping 3D points from the camera's coordinates system into 2D points on the image plane, and \(\mathbf{K}_k\) denotes the camera calibration matrix. In this study, we also follow \cite{brachmann2021visual, brachmann2023accelerated} to use the conditional rules for the re-projection error function based on the outcomes of predictions. Consequently, the training objective in Eq. \ref{eq:reprojection} is modified as follows:

\vspace{-1.5em}
\begin{align}
    \label{eq:condition_rule}
     r(\mathbf{x}_k,\mathbf{y}_k,h_i^*)=
     \begin{cases}
         \tau(t)\text{tanh}(\frac{\|\pi(\mathbf{K}_k{h_i^*}^{-1}\mathbf{y}_k)-\mathbf{x}_k\|}{\tau(t)}) & \text{if } \mathbf{y}_k \in \mathcal{V} \\
         \|\hat{\mathbf{y}_k} - \mathbf{y}_k\|  & \text{otherwise.}
     \end{cases}
\end{align}

Eq. \ref{eq:condition_rule} is designed to minimize the re-projection error \(\pi\) for all valid coordinates predictions \(\mathcal{V}\). These valid predictions must be located between 10cm and 1000m from the image plane and have a re-projection error of less than 1000px \cite{brachmann2021visual}. The threshold \(\tau\) is set in accordance with \cite{brachmann2023accelerated}:

\vspace{-1.5em}
\begin{align}
    \label{eq:tau}
    \tau(t) = w(t)\tau_{\mathrm{max}}+\tau_{\mathrm{min}}, \text{ with } w(t)=\sqrt{1-t^2},
\end{align}

where \(t \in (0,1)\) indicates the relative training progress. For invalid predictions in Eq. \ref{eq:condition_rule}, we reduce the distance between the heuristic targets \(\hat{\mathbf{y}_k}=h_i^*\hat{\mathbf{e}_k}\) and the re-projected dummy 3D coordinates \(\hat{\mathbf{e}_k}\), which are fixed at a depth of 10m. 

Once we have obtained the 2D-pixel positions \(\mathbf{x}_k\) and their corresponding predicted scene coordinates \(\mathbf{y}_k\), we can estimate the pose using a pose solver, as specified:

\vspace{-1.5em}
\begin{align}
    \label{eq:pose_solver}
    h_i = \zeta({\mathbf{x}_k,\mathbf{y}_k}).
\end{align}

The function \(\zeta(.)\) represents the pose solver. In this study, we use DSAC* \cite{brachmann2023accelerated} as our pose solver.

\begin{figure*}[t!]
\vskip -0.5pc
\begin{center}
\includegraphics[width=\textwidth]{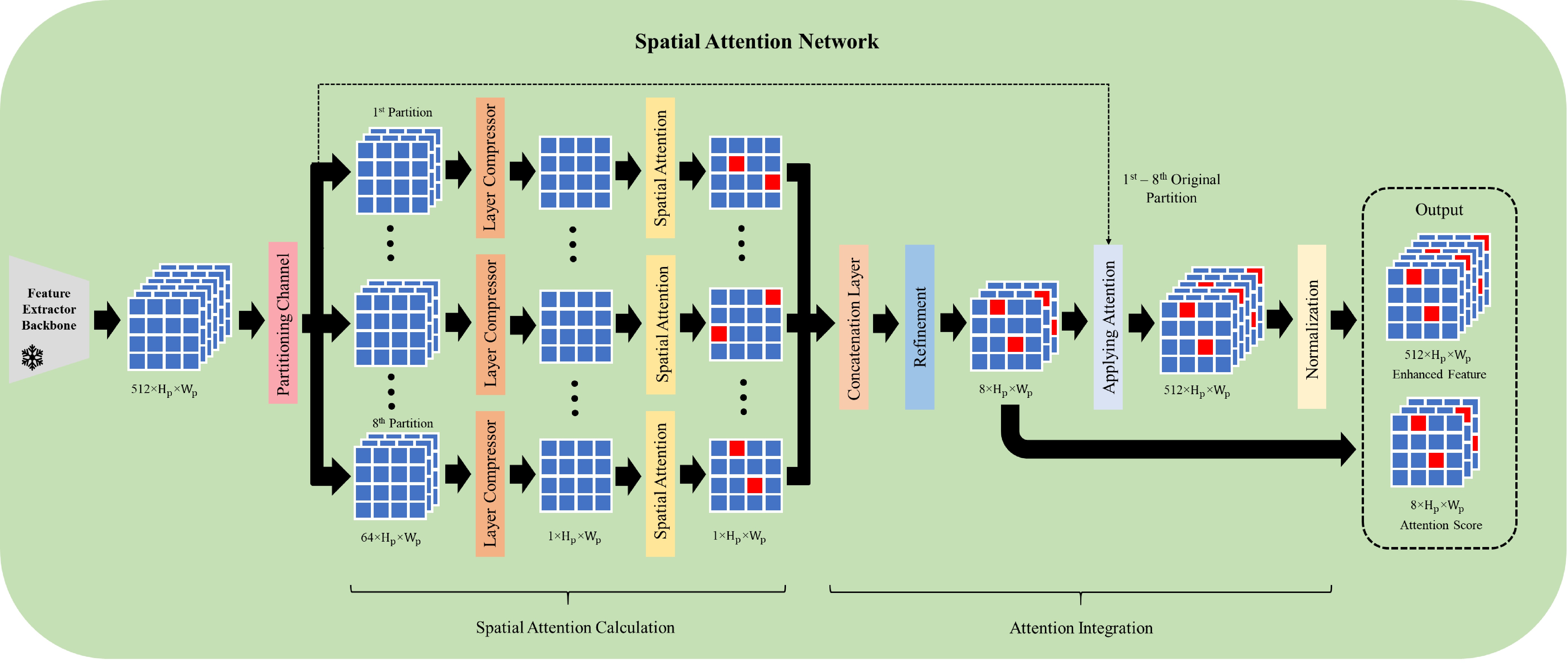}
\caption{\label{fig:spatial_attention}Overview of the Spatial Attention Network. Our spatial attention network contains two main parts: the spatial attention calculation and the attention integration. Using this spatial attention network, we choose the top 30\% features to be included in the training buffer.}

\end{center}
\vskip -1.5pc
\end{figure*}

\subsection{Important Feature Selection}
In addition to dividing the regression network as shown in Eq. \ref{eq:split_network} to reduce the training time, employing gradient decorrelation \cite{brachmann2023accelerated} will also speed up the convergence. The concept is straightforward: randomize the patches (essentially features) throughout the training set and build the training buffer using various mapping views. 

In the baseline's (ACE) framework, each extracted \(\mathfrak{f}_k\) comes with associated details such as pixel positions \(\mathbf{x}_k\), camera intrinsics \(\mathbf{K}_k\) and the ground truth mapping poses \(h_i^*\). These instances are typically selected randomly to compile the training buffer for network training. However, we argue that \(\mathfrak{f}_k\) selection for the training buffer should not be random but should instead focus on including only important instances (features). To achieve this, we employ a spatial attention network and an edge detector to identify and select these important instances for the training buffer. For a better understanding, we have included a visualization of the mapping process in Fig. \ref{fig:framework}.

In Fig. \ref{fig:framework}, we employ the outputs from the spatial attention network and the edge detector to select the important features to be included in the training buffer. First, the spatial attention network provides enhanced features and their corresponding attention scores. Utilizing this data, we select the top 30\% of the features and create a mask based on their locations, designating important regions where these top features are located. Similarly, for the results from the edge detector, we generate a mask based on the robustly detected edges, marking important regions where these edges occur. Once both masks are generated, we identify intersecting areas between them to define the final mask. However, in the experiment, we recognized that the mask generated by the edge detector tends to be smaller due to the limited extent of edges. Therefore, we also sample around the intersected mask within a specific radius, setting it to 7 pixels to avoid excessive restriction. Finally, we use the final mask to guide the selection of important features from those extracted by the backbone, which are then used to create the training buffer.

\subsection{Spatial Attention Network}
As depicted in Fig. \ref{fig:framework}, the spatial attention network is employed following the feature extraction by the backbone. Then, Fig. \ref{fig:spatial_attention} illustrates the proposed spatial attention network.

Our spatial attention model begins by dividing the input channels with dimensions \( (512 \times \mathrm{H_p} \times \mathrm{W_p}) \) into 8 partitions. Thus, each partition will have a dimension \( (64 \times \mathrm{H_p} \times \mathrm{W_p}) \). Consequently, we proceed with the spatial attention calculation and attention integration.

\textbf{Spatial Attention Calculation}. This process focuses on determining the spatial relationships for each previously partitioned channel. Each partitioned channel is compressed into a single channel using a convolutional layer with kernel size \(1 \times 1\). Once we have a single channel with dimension \((1 \times \mathrm{H_p} \times \mathrm{W_p})\), we proceed to compute spatial attention. We apply a convolutional layer with a \(7 \times 7\) kernel and padding 3, followed by a softmax function. This approach helps to identify and highlight areas within the patch that are likely to be of significant interest.

\textbf{Attention Integration}. At this section, we combine the attention maps from each partition to reconstruct the original feature dimensions. Initially, we concatenate the outputs of all partitions, followed by a refinement process using a convolutional layer with kernel size \(3 \times 3\). This refinement is applied to each of the 8 partitions separately via grouped convolution, resulting in an output dimension of \( (8 \times \mathrm{H_p} \times \mathrm{W_p}) \). This approach ensures that each partition refines its spatial data independently, improving the model’s ability to concentrate on specific features within each segment.

Consequently, we apply the refined attention back to the original partition inputs to restore the original dimensions \( (512 \times \mathrm{H_p} \times \mathrm{W_p}) \) and then perform normalization. Finally, the spatial network generates two outputs: the attention scores and the enhanced feature maps.

\begin{figure*}[t!]
\begin{center}
\includegraphics[width=0.8\textwidth]{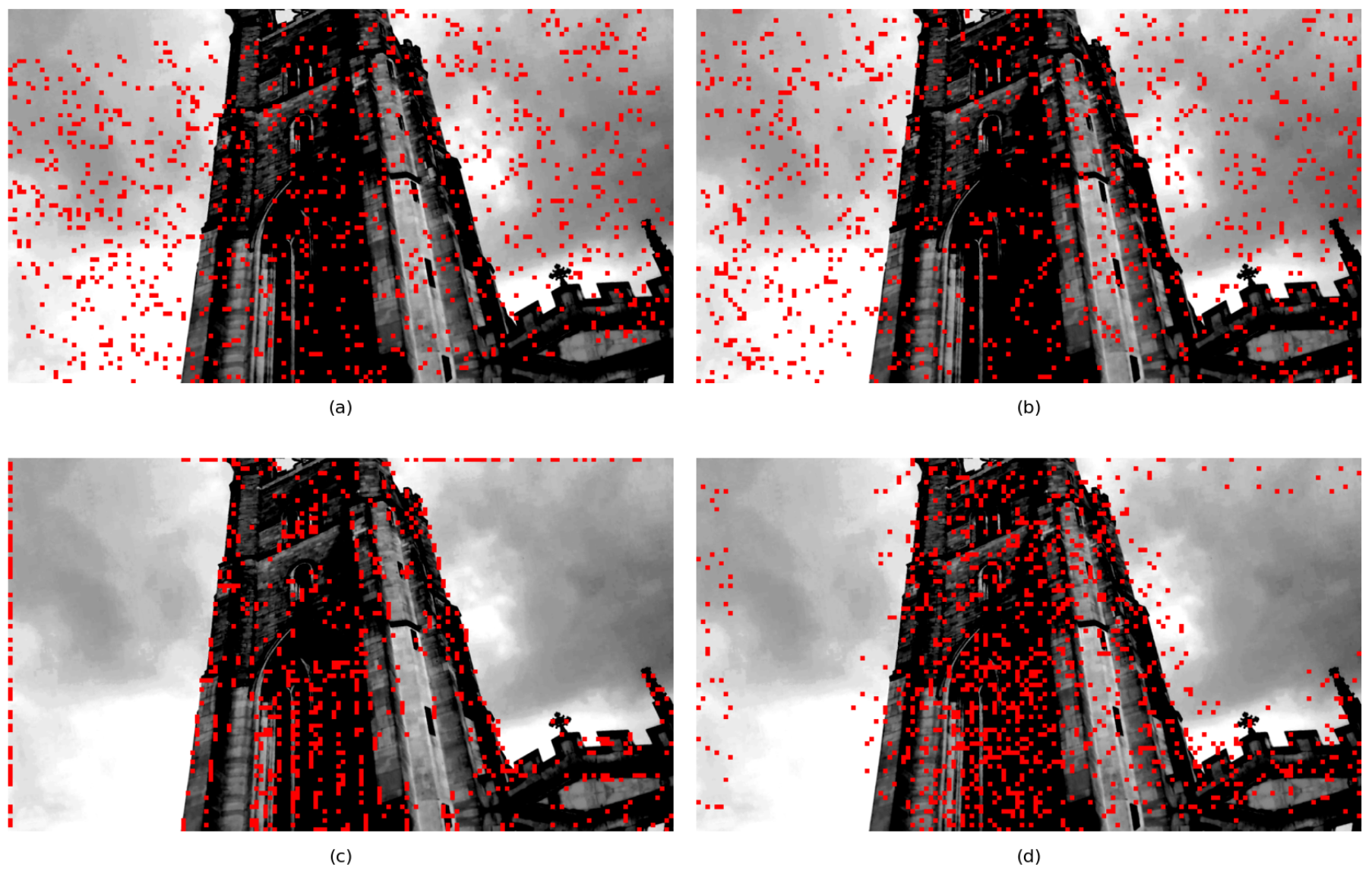}
\caption{\label{fig:comparison}Sampled features for the training buffer comparison. (a) ACE's sampled feature, (b) Only spatial attention network-based sampled feature, (c) Only edge detector-based sampled feature, (d) Spatial attention network and edge detector sampled feature. Using our proposed method (d), we avoid the featureless area that does not provide informative features for the network.}
\end{center}
\vskip -1.5pc
\end{figure*}

\begin{figure}[t!]
\begin{center}
\includegraphics[width=\columnwidth]{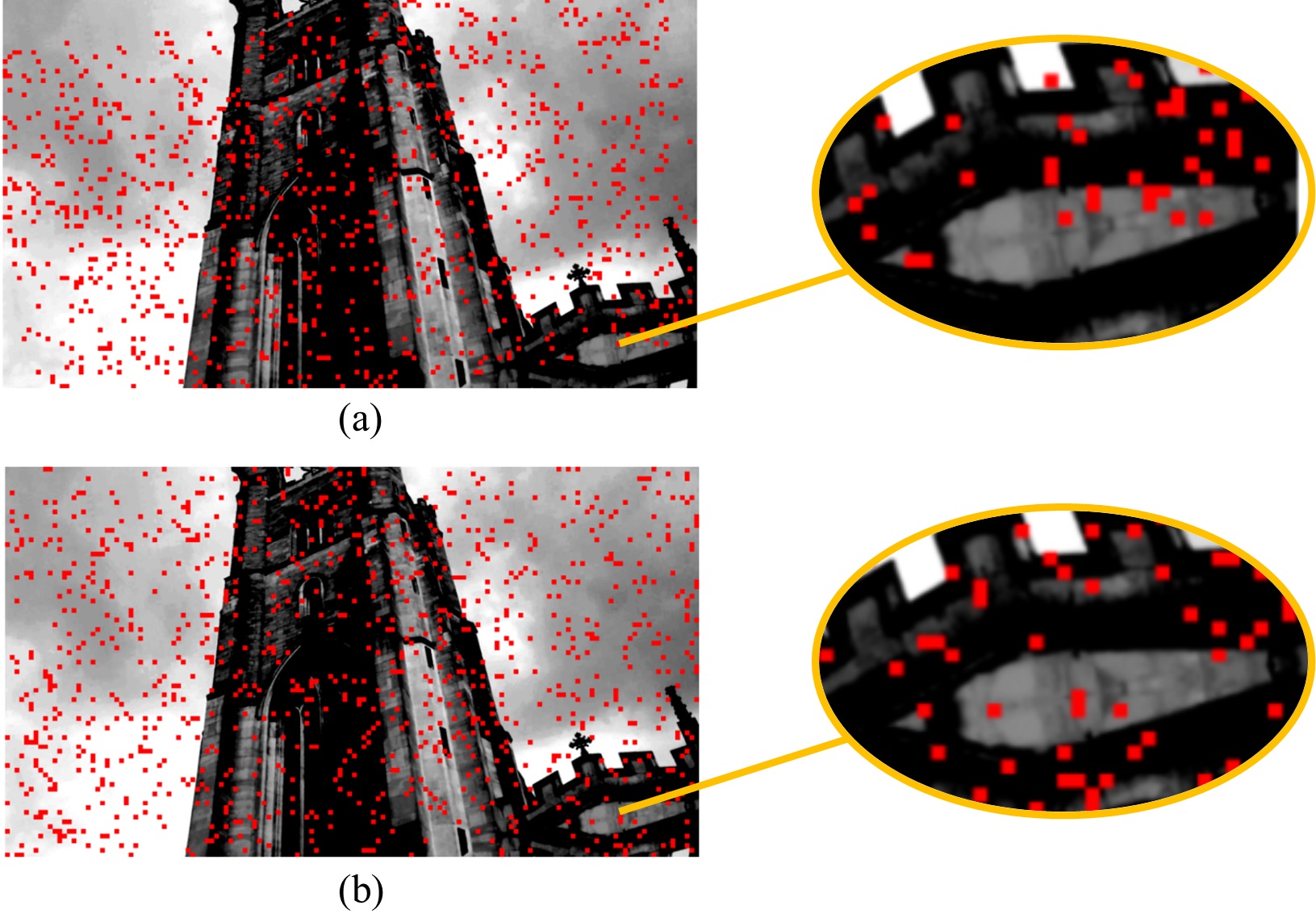}
\caption{\label{fig:comparison_attention}Comparison of ACE's sampled features and our spatial attention network's sampled features. (a) Features sampled using ACE's method, highlighting missed details. (b) Features identified exclusively by our spatial attention network, capturing areas overlooked by ACE.}

\end{center}
\vskip -1.5pc
\end{figure}

Figure \ref{fig:comparison}(b) showcases the results of the feature sampling conducted solely by our spatial attention network. This approach, while innovative, tends to select both informative and featureless regions of the image, which might be considered non-informative. This issue mostly comes from how our model slightly differs from the usual deep attention mechanisms. Normally, the deep attention mechanism uses complex calculations to better spotlight the key features. The simplification in our model, achieved by avoiding these complex mechanisms and reducing channel compression, unintentionally leads to some loss of information. However, this simplification is not without its merits. It allows for significantly faster mapping times and reduced mapping sizes, providing a practical advantage in scenarios where speed and efficiency are essential. Despite its drawbacks, this method shows modest improvements compared to ACE's sampling technique, as depicted in Fig. \ref{fig:comparison_attention}. It effectively identifies and highlights areas of the building shown in Fig. \ref{fig:comparison_attention}(b), which the ACE method overlooks in Fig. \ref{fig:comparison_attention}(a). This demonstrates its ability to capture important features that were previously missed.

\subsection{Edge Detector}
As \cite{lowe2004distinctive} states, the most reliable and robust keypoints are often located at the edges of scenes. To ensure the selection of reliable features for the training buffer, we incorporate an edge detection strategy by applying the Canny filter \cite{canny1986computational} to the mapping images \(I_M\). We consider these detected edges to be robust regions. As a result, we create a mask for these robust regions and proceed according to the feature selection steps outlined in Fig. \ref{fig:framework}. We visualize the sampled features by using the edge detector only in Fig. \ref{fig:comparison}(c), where it is clear that the edge detector effectively avoids featureless areas. However, we recognize that the edge detector's output can be overly restrictive, potentially increasing the risk of overfitting during network training.

Finally, as demonstrated in Fig. \ref{fig:comparison}(d), our approach, which combines the spatial attention network and the edge detector, successfully samples features from informative areas while steering clear of regions lacking distinctive characteristics.

\section{Experiment and Results}
We conducted a validation of our method using the publicly available ACE \cite{brachmann2023accelerated} PyTorch code and tested against the benchmark dataset, Cambridge Landmarks, and our custom dataset, JBNU. We tested the proposed method using NVIDIA RTX 4070Ti as our GPU. In alignment with the original settings of ACE, we maintained the same hyperparameters, except for the top class parameter $\sigma = 0.3$ and sampled radius $\gamma = 7$.  For detailed information on the other hyperparameters, please see \cite{brachmann2023accelerated}. 

\subsection{Dataset}

\begin{figure}[t!]
\vskip -0.5pc
\begin{center}
\includegraphics[width=8cm]{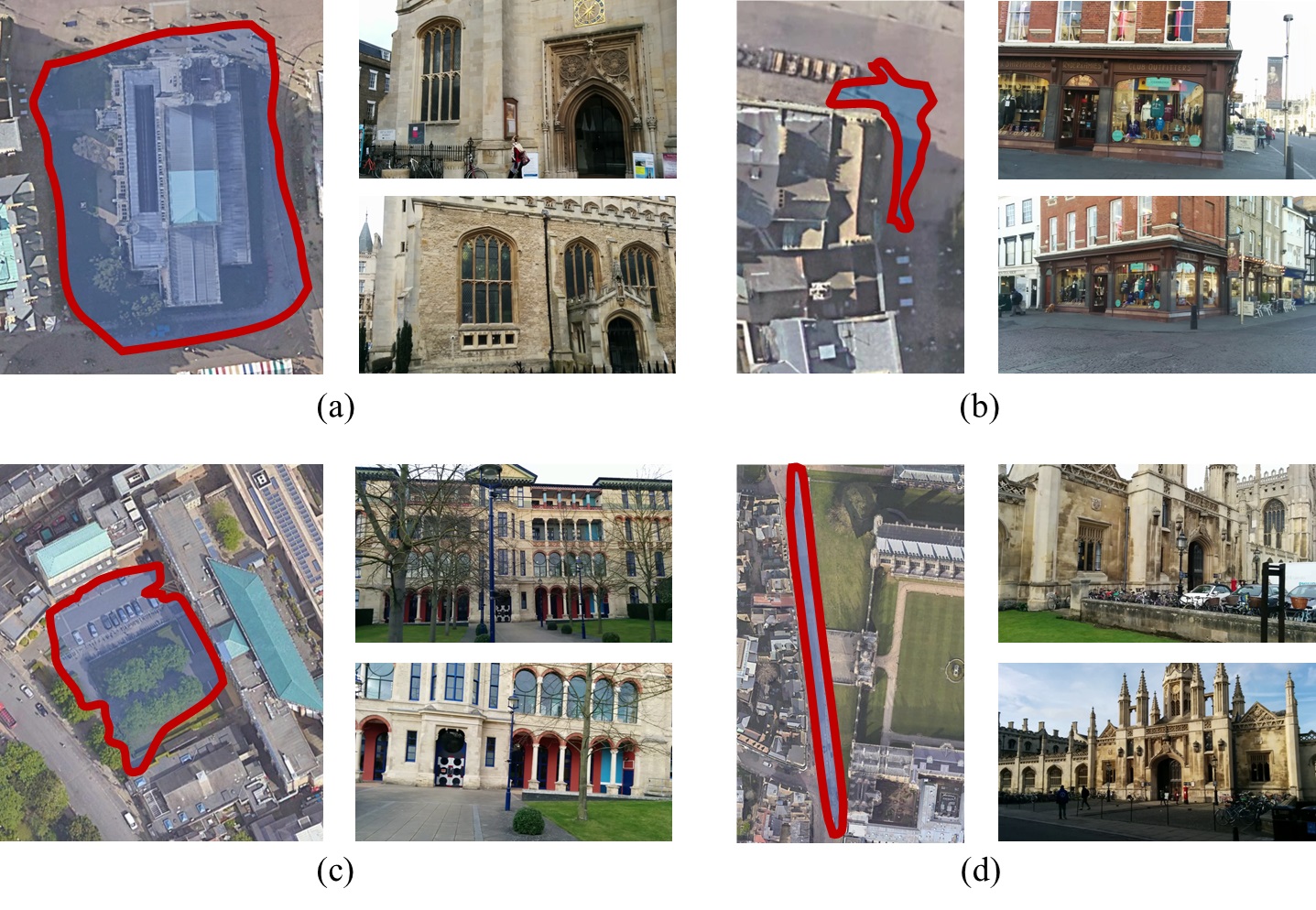}
\caption{\label{fig:aerial_cambridge}Cambridge Landmarks dataset. (a) St. Mary's Church, (b) Shop Facade, (c) Old Hospital, and (d) Kings College.}
\end{center}
\vskip -1.5pc
\end{figure}

\textbf{Cambridge Landmarks} \cite{kendall2015posenet}. This dataset contains five distinct sites within the University of Cambridge with the total distance covered by one scene being approximately 200 meters. When we accessed this dataset, we could only obtain data from four locations: St. Mary's Church, Shop Facade, Old Hospital, and Kings College. The ground truth poses for this dataset were established using VisualSFM \cite{wu2011visualsfm}. We notice that this dataset contains lots of pedestrians and vehicle scenes, making it a good challenge for the visual localization task. We visualize the Cambridge dataset in Fig. \ref{fig:aerial_cambridge}. The red-blue area is the region where the training and testing dataset was taken.

\begin{figure}[t!]
\vskip -0.5pc
\begin{center}
\includegraphics[width=8cm]{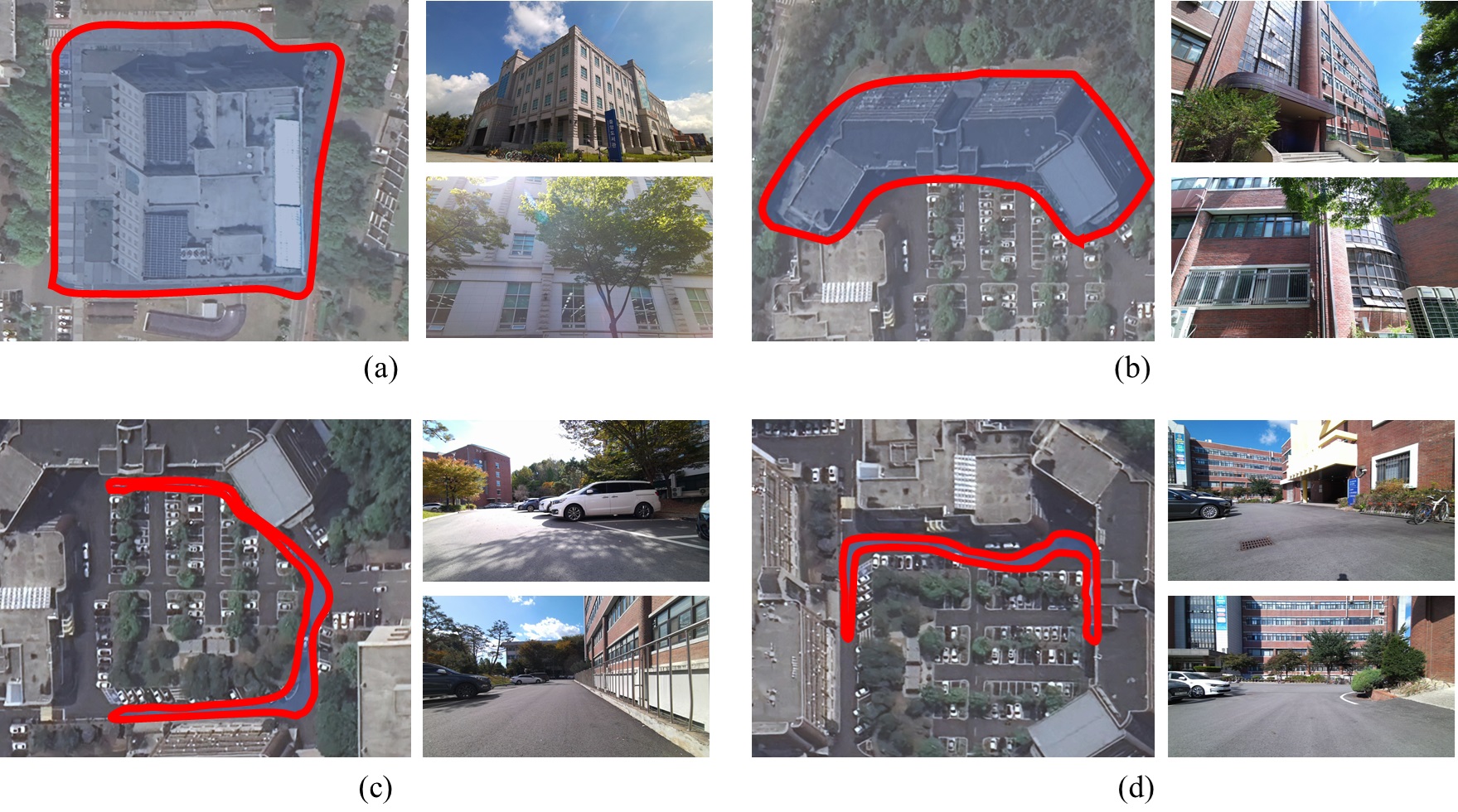}
\caption{\label{fig:aerial_jbnu}JBNU dataset. (a) Library, (b) Building 7, (c) Six-to-Seven, and (d) Seven-to-Six.}
\end{center}
\vskip -1.5pc
\end{figure}

\textbf{JBNU}. The JBNU dataset was created at Jeonbuk National University in South Korea. This custom dataset is classified as a large outdoor dataset due to its coverage of approximately 400 meters, which is twice the distance of the Cambridge Landmarks dataset. It includes four scenes: Library, Building 7, Six-to-Seven, and Seven-to-Six. We visualize the JBNU dataset in Fig. \ref{fig:aerial_jbnu}. The ground truth poses in real-world scale was produced using LIO-SLAM\cite{bai2022faster} and Structure from Motion (SfM)\cite{schonberger2016structure,schonberger2016pixelwise}. Similar to the Cambridge Landmarks dataset, the JBNU dataset presents challenges related to vehicle scenes as well as ambiguous scenes. The JBNU dataset consists of images, camera calibration, and its corresponding ground truth poses.

\subsection{Localization Results}

\begin{table}[t!]
    \caption{Comparison with the Cambridge Landmarks dataset. We report the percentage of test frames exhibiting a pose error of less than 5 meters and 10 degrees, noting that higher percentages indicate better accuracy. We also present the median rotation error (in degrees) and translation error (in centimeters) for the baseline method, ACE\cite{brachmann2023accelerated}, and our proposed method. A lower median error indicates superior performance. Across all scenes in the Cambridge Landmarks dataset, our method consistently outperforms ACE.}
    \label{tab:compare_ace}
    \centering
    \setlength{\tabcolsep}{3pt}
    \renewcommand{\arraystretch}{1.2}
    \small
    \begin{tabular}{lcccc} 
    \hline
    & \multicolumn{2}{c}{Accuracy (\%)} & \multicolumn{2}{c}{Median Error (° / cm)} \\
     & ACE & Ours & ACE & Ours \\ 
    \hline
    St. Mary & 82.3 & \textbf{88.3} & 0.7 / 22.528 & \textbf{0.5 / 17.394} \\
    Shop     & 100  & \textbf{100}  & 0.3 / 5.736 & \textbf{0.3 / 5.191} \\
    Hospital & 100  & \textbf{100}  & 0.6 / 29.111 & \textbf{0.5 / 27.406} \\
    College  & 98.8 & \textbf{99.4} & 0.4 / 28.122 & \textbf{0.4 / 24.730} \\ \hline
    \end{tabular}
\end{table}

\textbf{Cambridge Landmarks}. We began our evaluation by comparing our proposed method to the baseline network, ACE \cite{brachmann2023accelerated}, focusing on the Cambridge Landmarks dataset. We reported the percentage of test frames with a pose error less than 5 meters and 10 degrees, alongside median rotation and translation errors in degrees and centimeters, respectively. As detailed in Table \ref{tab:compare_ace}, our method not only surpassed ACE in terms of accuracy but also reduced median errors significantly across all scenes. Notably, accuracy in the St. Mary's Church scene improved by 6\%, from 82.3\% to 88.3\%. This enhancement is largely attributed to the integration of the spatial attention network and the edge detector that selectively focuses on robust features within the scene, contrasting with ACE’s random feature selection. Given the Cambridge dataset's distinct architectural styles and varying occlusions, such precise feature prioritization is crucial, especially in urban environments where the contextual relevance of features plays a key role in accurate localization.

\begin{table*}[!t]
  \caption{Localization results on Cambridge Landmarks. We report the median rotation error (in degrees) and translation error (in centimeters) for every method. In the SCR group, \textbf{bold} represents the best result, and \underline{underline} is the second-best result.}
  \label{tab:localized_cambridge_full}
  \centering
    \setlength{\tabcolsep}{3pt} 
    \renewcommand{\arraystretch}{1.2}
    \small
    \begin{tabular}{l l c c c c c c}
    \hline
     \multicolumn{2}{c}{\multirow{2}{*}{}} & \multicolumn{1}{c}{\multirow{2}{*}{\parbox{1.5cm}{\centering Mapping\\Time}}} & \multicolumn{1}{c}{\multirow{2}{*}{\parbox{1.5cm}{\centering Mapping\\Size}}} & \multicolumn{4}{c}{Cambridge} \\ \cline{5-8}
     & \multicolumn{1}{c}{} & \multicolumn{1}{c}{} & \multicolumn{1}{c}{} & \multicolumn{1}{c}{St. Mary} & \multicolumn{1}{c}{Shop} & \multicolumn{1}{c}{Hospital} & \multicolumn{1}{c}{College} \\ \hline
     \multirow{3}{*}{\parbox{1cm}{\centering FM}}
     & NetVLAD\cite{arandjelovic2016netvlad} + HLOC\cite{sarlin2019coarse}\cite{sarlin2020superglue} & \multirow{3}{*}{\parbox{1.5cm}{\centering $\sim$1.5h}}   &  $\sim$1.8GB   &     0.3 / 8.9   &      0.2 / 4.2   &   0.4 / 20.3      &  0.2 / 13.1       \\ 
     & APGeM\cite{revaud2019learning} + R2D2\cite{revaud2019r2d2}   &   &  $\sim$2.2GB   & 0.3 / 9.9    &  0.2 / 4.1   & 0.3 / 17.5     &  0.2 / 12.7   \\ 
     & pixLoc\cite{sarlin2021back}                                  &   &  $\sim$1GB     &     0.3 / 10      &      0.2 / 5      &    0.3 / 16       &   0.2 / 14           \\ \hline
     \multirow{2}{*}{\parbox{1cm}{\centering APR}} 
     & PoseNet\cite{kendall2017geometric}      & $\sim$24h  &  $\sim$50MB   &     3.3 / 157    &      3.8 / 88   &    3.3 / 320     &   1.0 / 88         \\ 
     & MS-Transformer\cite{shavit2021learning} & $\sim$7h  &   $\sim$18MB   &     4.0 / 162     &      3.1 / 86   &    2.4 / 181       &   1.5 / 83           \\ \hline
     \multirow{3}{*}{\parbox{1cm}{\centering SCR}}
     & DSAC*\cite{brachmann2021visual}    & 15.5h   &  27.5MB   &  \textbf{0.4 / 12.3}     &  \textbf{0.2 / 5.1}      &   \textbf{0.5 / 21.9}     &   \textbf{0.3 / 15.7}   \\ 
     & ACE\cite{brachmann2023accelerated} & 255s  &  4.2MB    &  0.7 / 22.5                       &  0.3 / 5.7                        &   0.6 / 29.1                       &    0.4 / 28.1          \\ 
     & Ours                               & 242s  &  5.3MB    &  \underline{0.5 / 17.4}   & \underline{0.3 / 5.2}    &   \underline{0.5 / 27.4}     &    \underline{0.4 / 24.7}          \\ \hline
  \end{tabular}
\end{table*}

Subsequently, we extended our analysis to include comparisons using feature matching, APR, and SCR-based methods, as shown in Table \ref{tab:localized_cambridge_full}. This comprehensive comparison is designed to evaluate the efficacy of our approach relative to the established feature matching-based method and to compare it with other deep learning-based methods such as APR. Specifically, APR uses a straightforward method where neural networks directly regress the camera pose, while SCR initially identifies 2D-3D correspondences with neural networks, which are then used to determine the camera pose using a pose solver.

According to Table \ref{tab:localized_cambridge_full}, the feature matching (FM) method remains exceptionally effective, particularly benefiting from image retrieval integrations, which enhance their localization precision. We can see if the FM's translation error achieves superior results while maintaining commendable rotation accuracy. This advantage probably stems from FM's capability to capture and leverage high-quality features across different settings. For example, our implementation utilizes SuperPoint \cite{detone2018superpoint} and R2D2 \cite{revaud2019r2d2}, which are famous for their strong feature detection abilities that are crucial for precise translation. However, FM methods are resource-intensive, requiring up to 2.2GB for mapping and about 1.5 hours to process, which poses significant challenges in dynamic environments where rapid processing is crucial. 

In contrast, the simpler approach employed by APR generates inferior outcomes compared to other techniques, highlighting the compromise between the complexity of the method and the accuracy of localization. APR’s approach, which relies on a single neural network pipeline for direct pose regression, significantly contributes to its performance limitations. This strategy limits the method's capacity to adjust dynamically to different environmental complexities and to independently optimize for various stages of the localization process, such as feature detection and pose calculation, similar to the feature matching method or the SCR method.

In Table \ref{tab:localized_cambridge_full}, within the SCR category, our method ranked second, only outperformed by DSAC*. It is crucial to note that while DSAC* delivered the best performance in terms of accuracy on the Cambridge dataset, it required about 15 hours of processing on a high-performance GPU. Our method, however, completed similar tasks in just 242s on the Cambridge dataset, highlighting not only our method's efficiency but also its capability to deliver competitive accuracy with drastically reduced mapping time. In terms of mapping size, our proposed method also has a reasonably good mapping size compared to the DSAC* and ACE. 

\begin{figure}[t!]
\vskip -0.5pc
\begin{center}
\includegraphics[width=8cm]{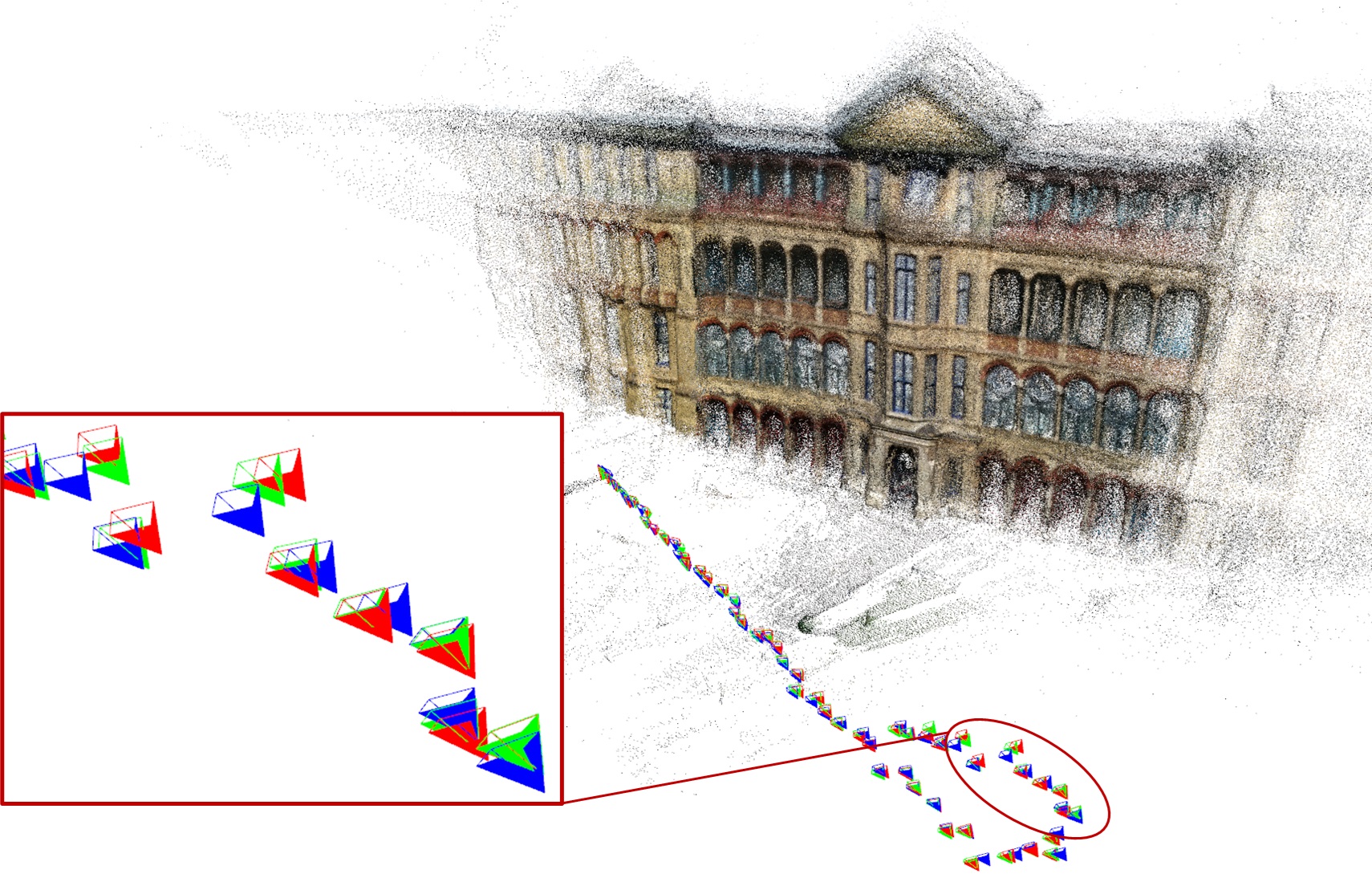}
\caption{\label{fig:hosiptal_result}Localization result of Old Hospital Cambridge Landmarks. Blue represents the ground truth, red indicates ACE, and green denotes our method. We demonstrate that our method exhibits lower translational and rotational errors compared to ACE.}
\end{center}
\vskip -1.5pc
\end{figure}

We also provide a visualization comparing the localization performance of our proposed method with ACE in the Old Hospital scene in Fig. \ref{fig:hosiptal_result}.

\begin{table*}[!t]
  \caption{Localization results on JBNU. We report the median rotation error (in degrees) and translation error (in meters) for every method. Additionally, we implement an ensemble model for potential models within the SCR group. In the SCR and ensemble group, \textbf{bold} represents the best result, and \underline{underline} is the second-best result.}
  \label{tab:localized_jbnu}
  \centering
    \setlength{\tabcolsep}{3pt} 
    \renewcommand{\arraystretch}{1.2}
    \small
    \begin{tabular}{l l c c c c c}
    \hline
     \multicolumn{2}{c}{\multirow{2}{*}{}} & \multicolumn{1}{c}{\multirow{2}{*}{\parbox{1.5cm}{\centering Mapping\\Size}}} & \multicolumn{4}{c}{JBNU} \\ \cline{4-7}
     & \multicolumn{1}{c}{} & \multicolumn{1}{c}{} & \multicolumn{1}{c}{Library} & \multicolumn{1}{c}{Building 7} & \multicolumn{1}{c}{Six-to-Seven} & \multicolumn{1}{c}{Seven-to-Six} \\ \hline
     \multirow{2}{*}{\parbox{1.5cm}{\centering FM}} 
     & NetVLAD\cite{arandjelovic2016netvlad} + R2D2\cite{revaud2019r2d2}                              &  $\sim$2.1GB   &  1.1 / 2.319   & 2.8 / 1.091  &  2.4 / 1.800 & 0.9 / 0.892     \\ 
     & APGeM\cite{revaud2019learning} + HLOC\cite{sarlin2019coarse}\cite{sarlin2020superglue}    &  $\sim$2.8GB   & 1.1 / 2.318  &  2.7 / 1.105 &  2.1 / 1.646  & 1.0 / 0.890   \\ \hline
     \multirow{4}{*}{\parbox{1.5cm}{\centering SCR}}
     & DSAC*\cite{brachmann2021visual}                           &  27.5MB   &  1.9 / 5.001             &  \textbf{3.3 / 1.632}       &   2.2 / 2.665               &    1.3 / 0.953   \\ 
     & ACE\cite{brachmann2023accelerated}                        &  4.2MB    &  \underline{2.0 / 3.933} &  7.8 / 2.287                &   \underline{1.9 / 2.009}   &    \underline{1.1 / 0.732}    \\ 
     & FocusTune (with dummy 3D map) \cite{nguyen2024focustune}  &  4.2MB    &  2.8 / 5.839             &  \underline{7.1 / 2.504}    &   2.2 / 1.694               &    1.1 / 0.818    \\
     & Ours                                                      &  5.3MB    &  \textbf{2.0 / 3.852}    &  8.2 / 2.889                &   \textbf{1.7 / 1.899}      &    \textbf{1.0 / 0.657}    \\ \hline
     \multirow{3}{*}{\parbox{1.5cm}{\centering Ensemble}}
     & ACE\cite{brachmann2023accelerated}                        &  16.8MB    &  \underline{0.6 / 2.151}   & \underline{2.5 / 1.040}     &   1.5 / 1.825                & \underline{1.1 / 0.600}   \\ 
     & FocusTune (with dummy 3D map) \cite{nguyen2024focustune}  &  16.8MB    &  1.2 / 2.104               &  2.9 / 1.061                &  \underline{1.5 / 1.760}     &    1.1 / 0.666            \\ 
     & Ours                                                      &  21.2MB    &  \textbf{0.6 / 2.095}      &  \textbf{2.2 / 1.038}       &  \textbf{1.4 / 1.809}        &  \textbf{1.1 / 0.598}      \\ \hline
  \end{tabular}
\end{table*}

\textbf{JBNU}. The JBNU dataset, developed at Jeonbuk National University in South Korea, represents a significant challenge due to its large outdoor coverage, spanning approximately 400 meters—twice the distance covered by the Cambridge Landmarks dataset. This dataset includes 4 scenes that present unique navigational challenges. Because the JBNU environment consists of ambiguous scenes—similar landmarks in several frames, accurately finding locations within this dataset is particularly difficult.

In Table \ref{tab:localized_jbnu}, feature matching (FM) techniques demonstrate their strengths in the JBNU dataset, particularly standing out in terms of translation accuracy while providing consistent rotation metrics. Similar to the Cambridge dataset, advanced algorithms like SuperPoint and R2D2 are utilized, reinforcing FM's ability to extract robust features from the ambiguous scene. However, FM's significant computational demands—requiring mapping sizes of up to 2.8GB and extensive processing times—become even more evident in this large and complex setting. 

In the SCR group, our method achieved impressive results, surpassing other competitors such as DSAC*. However, it's important to note that DSAC*, while showing the best performance in the more uniform and predictable Cambridge dataset, struggles in the larger and ambiguous dataset like JBNU. This drop in performance is due to the frequent occurrence of similar architectural elements across different frames in JBNU, which complicates the task of precise feature localization. This issue limits the effectiveness of DSAC*. In contrast, our approach leverages a spatial attention network and an edge detector that selectively processes only the crucial features. This approach helps to avoid confusion caused by ambiguous elements, enabling our system to maintain high accuracy with considerably lower computational demands. This is also shown in Table \ref{tab:localized_jbnu}, where our method achieves the lowest median error across the majority of JBNU scenes.

\begin{figure}[t!]
\vskip -0.5pc
\begin{center}
\includegraphics[width=8cm]{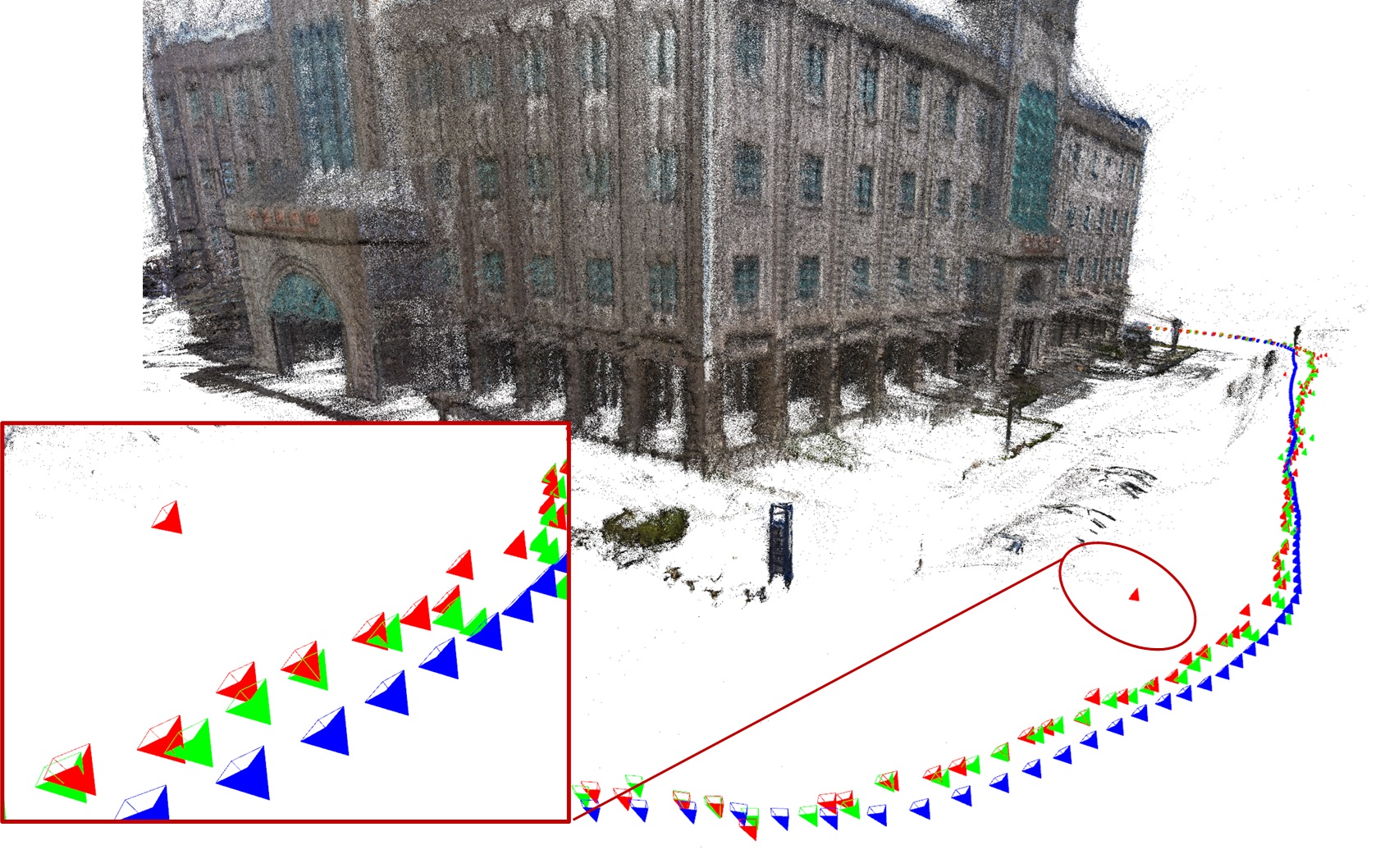}
\caption{\label{fig:library_result}Localization result of Library JBNU. Blue represents the ground truth, red indicates ACE, and green denotes our method. Each method is integrated into an ensemble model. We show that our approach follows ground truth pose well with minimum error compared  to ACE.}
\end{center}
\vskip -1.5pc
\end{figure}

In the JBNU dataset experiment, we also employed the ensemble model experiment. The concept of the ensemble model was initially derived from our baseline ACE. In an extensive area, a single scene coordinate regression network often fails to cover the entire domain \cite{brachmann2019expert}. To address this, the ensemble model divides the mapping frames into sub-regions, with multiple regression heads assigned to each. These heads concentrate on specific scene areas. For instance, in our study, we divided the scene into four sub-regions and assigned four heads, training each separately using our proposed network. Furthermore, we implemented the ensemble model approach in other relevant networks such as ACE and FocusTune.

The results of the ensemble model demonstrate an improvement over the localization outcomes of the SCR group, suggesting that segmenting a large mapping area into smaller regions enhances the effectiveness of intensive training regimens. However, in datasets with naturally smaller mapping areas, like the Seven-to-Six scene, this strategy does not offer substantial benefits, suggesting that segmenting an already small scene is not advantageous.

As shown in Table \ref{tab:localized_jbnu}, the ensemble model also exceeds the median rotation error results achieved by the FM method. Specifically, in the Building 7 scene, employing the ensemble model with our proposed network reduced the rotation error from 2.7° to 2.2°. The outcomes of the ensemble model in the Library scene are visually presented in Fig. \ref{fig:library_result}.

\subsection{Ablation Study}

We perform an ablation study to evaluate the impact of variations in the top class parameter $\sigma$ and the sampled radius $\gamma$ on localization accuracy, as detailed in Table \ref{tab:class_variance}. The parameter $\sigma$ indicates the percentage of the highest scoring features to be sampled; for instance, $\sigma=0.1$ implies that only the top $10\%$ of scoring features are selected. The parameter $\gamma$ defines the sampling radius; for example, $\gamma=5$ means features within a 5-pixel radius of identified informative features are also sampled.

\begin{table}[!h]
    \caption{Ranking and radius variance. The variable $\sigma$ denotes the percentage of top-class samples, while $\gamma$ signifies the radius employed for sampling. We perform the localization upon Cambridge Landmarks: St. Mary's Church. We report the median rotational error (in degrees) and translation error (in cm). The best result is highlighted in \textbf{bold}.}
    \label{tab:class_variance}
    \centering
    \setlength{\tabcolsep}{3pt}
    \renewcommand{\arraystretch}{1.2}
    \small
    \begin{tabular}{cccccc}
    \hline
        \multicolumn{6}{c}{St. Mary's Scene} \\ \hline
        \diagbox[width=3.2em,height=2em]{$\gamma$}{$\sigma$} & $0.1$ & $0.2$ & $0.3$ & $0.4$ & $0.5$ \\ \hline
        $5$    &   0.7 / 21.8   &   0.7 / 23.8   &   0.6 / 21.1            &  0.7 / 21.9  &   0.7 / 23.4 \\
        $7$    &   0.7 / 22.5   &   0.6 / 21.3   &   \textbf{0.5 / 17.4}   &  0.7 / 21.9  &   0.5 / 18.9 \\ \hline
    \end{tabular}   
\end{table}

Table \ref{tab:class_variance} illustrates that using a stricter top class parameter $\sigma$ does not necessarily improve results. Similarly, increasing the sampled radius $\gamma$ does not guarantee better localization outcomes. Additionally, it is important to note that setting the top class parameter $\sigma$ too high, such as above $50\%$, leads to outcomes comparable to random sampling, as seen in ACE \cite{brachmann2023accelerated}, due to the majority of features being treated equivalently.

\section{Conclusion}
In this paper, we introduced a simple spatial attention network designed to select the informative features within the training buffer of the scene regression network. To further refine the selection process and eliminate outliers, we integrated an edge detector, ensuring that the chosen features are located in robust regions. Our results demonstrate that leveraging spatial relationships between patch information along the detected edges significantly enhances localization accuracy.

We integrated this approach into the ACE network, achieving significant reductions in both translational and rotational errors. Looking ahead, we intend to develop a more focused spatial attention network—without supplementary modules like the edge detector—to handle localization tasks in large and complex datasets.


\bibliographystyle{unsrt}
\bibliography{manuscript}

\relax 

\end{document}